\documentclass[runningheads]{llncs}
\usepackage{graphicx}
\usepackage{tikz}
\usepackage{comment}
\usepackage{amsmath,amssymb} 
\usepackage{color}
\usepackage{multirow}
\usepackage{makecell}
\usepackage[caption = false]{subfig}
\usepackage{floatrow}
\newfloatcommand{capbtabbox}{table}[][\FBwidth]
\usepackage{microtype}
\usepackage{caption}
\usepackage{caption}
\usepackage{booktabs}

\newcommand\bb[1]{\textbf{#1}}
\newcommand\R{\mathbb{R}}

\newcommand{\pms}[1]{{\scriptsize $\pm$#1}}

\begin{document}
\pagestyle{headings}
\mainmatter
\def\ECCVSubNumber{1267}  

\title{Selecting Relevant Features from a Multi-domain Representation for Few-shot Classification} 

\titlerunning{Selecting Relevant Features from a Multi-domain Representation for FSL}
%
\author{Nikita Dvornik\inst{1} \and
Cordelia Schmid\inst{2} \and
Julien Mairal\inst{1}}
\authorrunning{N. Dvornik et al.}
%
\institute{Univ. Grenoble Alpes, Inria, CNRS, Grenoble INP, LJK, 38000 Grenoble,
  France \and
Inria, \'Ecole normale sup\'erieure, CNRS, PSL Research Univ., 75005 Paris, France
\email{firstname.lastname@inria.fr}}
\maketitle

\begin{abstract}
   Popular approaches for few-shot classification consist of first learning a
generic data representation based on a large annotated dataset, before
adapting the representation to new classes given only a few labeled samples. In
this work, we propose a new strategy based on feature selection, which is both
simpler and more effective than previous feature adaptation approaches. First,
we obtain a multi-domain representation by training a set of semantically different
feature extractors. Then, given a few-shot learning task, we use our multi-domain
feature bank to automatically select the most relevant representations. We show
that a simple non-parametric classifier built on top of such features produces
high accuracy and generalizes to domains never seen during training, leading
to state-of-the-art results on MetaDataset and improved accuracy on
\textit{mini}-ImageNet.


\end{abstract}

\section{Introduction}
Convolutional neural networks~\cite{lecun1989backpropagation} (CNNs) have become
a classical tool for modeling visual data and are commonly used in
many computer vision tasks such as image
classification~\cite{krizhevsky2012imagenet}, object detection
\cite{blitznet,ssd,faster-rcnn}, or semantic segmentation
\cite{blitznet,long2015fully,ronneberger2015u}. One key of the success of 
these approaches relies on 
massively labeled datasets
such as ImageNet~\cite{imagenet} or COCO~\cite{coco}.
Unfortunately, annotating data at this scale is expensive and not always
feasible, depending on the task at hand. Improving the generalization
capabilities of deep neural networks and removing the need for huge sets of
annotations is thus of utmost importance.

This ambitious challenge may be addressed from different perspectives, such as
large-scale unsupervised learning~\cite{caron2018deep}, self-supervised
learning~\cite{doersch2017multi,gidaris2018unsupervised}, or by developing
regularization techniques dedicated to deep
networks~\cite{bietti2019kernel,yoshida2017spectral}. An alternative solution is
to use data that has been previously annotated for a different task than the one considered,
for which only a few annotated samples may be available.
This approach is particularly
useful if the additional data is related to
the new task~\cite{yosinski2014transferable,vtab}, which is unfortunately not known beforehand.
How to use effectively this additional data is then an important subject of ongoing
research~\cite{finn2017model,triantafillou2019meta,vtab}. In this paper, we
propose to use a multi-domain image representation, \textit{i.e.}, an
exhaustive set of semantically different features. Then, by automatically selecting only relevant
feature subsets from the multi-domain representation, we show 
how to successfully solve a large variety of target tasks.

\begin{figure}[t]
   \centering
  \includegraphics[width=\linewidth,trim=0 0 0 0,clip]{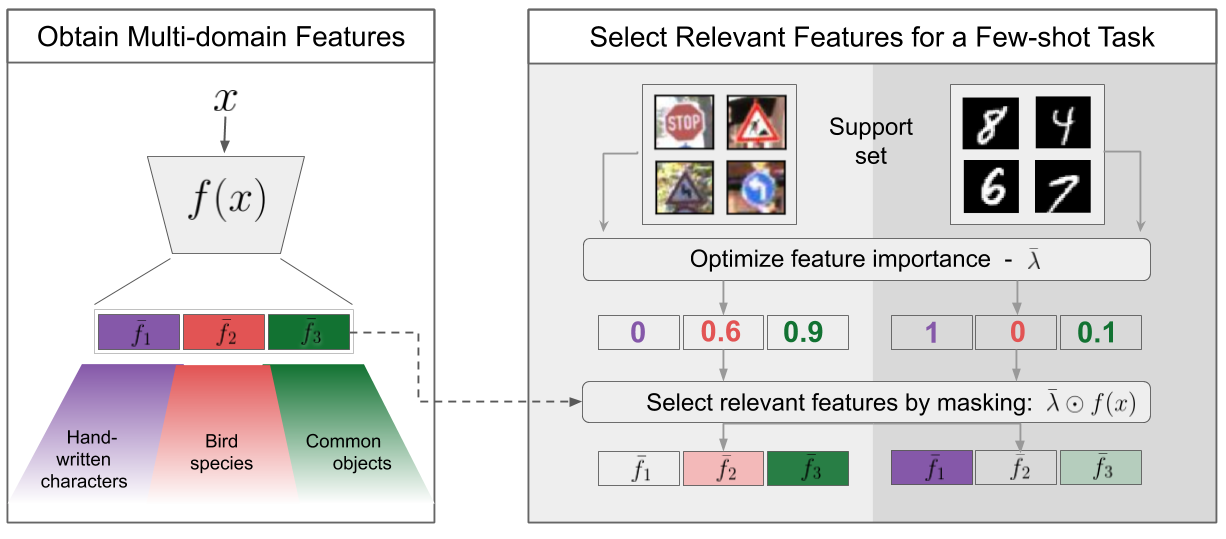}
\caption{\textbf{Illustration of our approach.} (Left) First, we obtain a
  multi-domain feature representation, consisting of feature blocks with different
  semantics. (Right) Given a few-shot task, we select only the relevant
  feature blocks from the multi-domain representation, by optimizing masking
  parameters $\lambda$ on the support set.}\label{fig:sketch_relationships}
\end{figure}

Specifically, we are interested in few-shot classification, where a visual model
is first trained from scratch, \textit{i.e.} starting from randomly initialized weights,
using a large annotated corpus. Then, we evaluate its ability to transfer the
knowledge to new classes, for which only very few annotated samples are
provided. Simply fine-tuning a convolutional neural network on a new
classification task has been shown to perform poorly~\cite{finn2017model}. This
has motivated the community to develop dedicated techniques, allowing effective
adaptation with few samples.

Few-shot classification methods typically operate in two stages, consisting of 
first pre-training a general feature extractor and then building an adaptation
mechanism. A common way to proceed is
based on
meta-learning~\cite{finn2017model,ravioptimization,schmidhuber1997shifting,snell2017prototypical,thrun1998lifelong,vinyals2016matching},
which is a principle to learn how to adapt to new learning
problems. That is, a parametric adaptation module (typically, a neural network)
is trained to produce a classifier for new categories, given only a few
annotated samples~\cite{gidaris2018dynamic,qiao2018few,rusu2018meta}.
To achieve this goal, the large training corpus is split into smaller
few-shot classification
problems~\cite{finn2017model,ravioptimization,vinyals2016matching}, that are
used to train the adaptation mechanism. Training
in such episodic fashion is advocated to alleviate overfitting and
improve
generalization~\cite{finn2017model,ravioptimization,vinyals2016matching}.
However, a more recent work~\cite{chen19closerfewshot} demonstrates that using
large training set to train the adapter is not necessary, \textit{i.e.} a linear
classifier with similar accuracy can be trained directly from new samples, on
top of a fixed feature extractor. Finally, it has been shown that 
adaptation is not necessary at all~\cite{dvornik2019diversity}; using a
non-parametric prototypical classifier~\cite{mensink2013distance} combined with a
properly regularized feature extractor can achieve better
accuracy than recent meta-learning baselines~\cite{dvornik2019diversity,saikia2020optimized}. These results suggest
that on standard few-shot learning
benchmarks~\cite{ravioptimization,ren2018meta},
the little amount of samples in a few-shot task is not enough to learn a
meaningful adaptation strategy.

To address the shortcomings of existing few-shot benchmarks, the authors of
\cite{triantafillou2019meta} have proposed MetaDataset, which evaluates the
ability to learn across different visual domains and to generalize to new data
distributions at test time, given few annotated samples. While methods based
solely on pre-trained feature extractors~\cite{saikia2020optimized} can
achieve good results only on test datasets that are similar to the
training ones, the adaptation
technique~\cite{requeima2019fast} performs well across test domains. The
method not only predicts a new few-shot classifier but also adapts
the filters of a feature extractor, given an input task. The results thus
suggest that feature adaptation may be in fact useful to achieve better generalization.

In contrast to these earlier approaches, we show that feature adaptation can be replaced by 
a simple feature selection mechanism, leading to better results  in the
cross-domain setup of~\cite{triantafillou2019meta}.
More precisely, we propose to
leverage a multi-domain representation -- a large set of semantically different features
that captures different modalities of a training set. Rather than adapting existing
features to a new few-shot task, we propose to select features from the
multi-domain representation. We call our approach SUR which stands for Selecting
from Universal Representations. To be more clear, we say universal because SUR
could be applied not only to multi-domain representations, but to any set of
representations that are semantically different. In contrast to standard adaptation
modules~\cite{oreshkin2018tadam,requeima2019fast,triantafillou2019meta} learned
on the training set, selection is performed directly on new few-shot tasks using
gradient descent.
Approaching few-shot learning with SUR has several advantages over
classical adaptation techniques. First, it is simple by nature, \textit{i.e.} selecting
features from a fixed set is an easier problem than learning a feature
transformation, especially when few annotated images are available. Second,
learning an adaptation module on the meta-training set is likely to generalize
only to similar domains.
In contrast, the selection step in our approach is decoupled from
meta-training, thus, it works equally well for any new domain.
Finally, we show that our approach achieves better results than current
state-of-the-art methods on popular few-shot learning benchmarks. In summary,
this work makes the following contributions:
\begin{itemize}
  \item We propose to tackle few-shot classification by selecting relevant
    features from a multi-domain representation. While multi-domain representations
    can be built by training several feature extractors or using a single
    neural network, the selection procedure is implemented with gradient descent.
  \item We show that our method outperforms existing approaches in in-domain and
    cross-domain few-shot learning and sets new state-of-the-art result on
    MetaDataset~\cite{triantafillou2019meta} and improves accuracy on
    \textit{mini}-ImageNet~\cite{ravioptimization}.
\end{itemize}
Our implementation is available at \url{https://github.com/dvornikita/SUR}.

\section{Related Work}
In this section, we now present previous work on few-shot classification and
multi-domain~representations, which is a term first introduced in~\cite{bilen2017universal}.

\subsubsection{Few-shot classification.}
Typical few-shot classification problems consist of two parts called
meta-training and meta-testing~\cite{chen19closerfewshot}. During the
meta-training stage, one is given a large-enough annotated dataset, which is
used to train a predictive model. During meta-testing, novel categories are
provided along with few annotated examples. The goal is to evaluate the ability
of the predictive model to adapt and perform well on these new classes.

Typical few-shot learning
algorithms~\cite{gidaris2019boosting,gidaris2018dynamic,qiao2018few} first
pre-train the feature extractor by supervised learning on the
meta-training set. Then, they use
meta-learning~\cite{schmidhuber1997shifting,thrun1998lifelong} to train an
adaptation mechanism. 
For example, in~\cite{gidaris2018dynamic,qiao2018few}, adaptation consists of
predicting the weights of a classifier for new categories,
given a small few-shot
training set. 
The work of~\cite{oreshkin2018tadam} goes beyond
the adaptation of a single layer on top of a fixed feature extractor, and additionally
generates FiLM~\cite{perez2018film} layers that modify convolutional layers.
Alternatively, the work of~\cite{chen19closerfewshot} proposes to train a linear
classifier on top of the features directly from few samples from new categories.
In the same line of work, \cite{lifchitz2019dense} performs implanting, \textit{i.e.}
learning new convolutional filters within the existing CNN
layers.

Other methods do not perform adaptation at all. It has been
shown in~\cite{dvornik2019diversity,lifchitz2019dense,saikia2020optimized} that
training a regularized CNN for classification on the meta-training set
and using these features directly with a nearest centroid classifier produces
state-of-the-art few-shot accuracy. To obtain a robust feature extractor,
\cite{dvornik2019diversity} distills an ensemble of networks into a single extractor
to obtain low-variance features. 


Finally, the
methods~\cite{requeima2019fast,saikia2020optimized} are the most relevant to our
work as they also tackle the problem of cross-domain few-shot
classification~\cite{triantafillou2019meta}. In~\cite{requeima2019fast},
the authors propose to
adapt each hidden layer of a feature extractor for a new task. They first obtain
a task embedding and use conditional neural
process~\cite{garnelo2018conditional} to generate parameters of modulation FiLM~\cite{perez2018film} layers,
as well as weights of a classifier for new categories.
An adaptation-free method~\cite{saikia2020optimized} instead trains a CNN on
ImageNet, while optimizing for high validation accuracy on other datasets, using
hyper-parameter search. When tested on domains similar to ImageNet, the method
demonstrates the highest accuracy, however, it is outperformed by
adaptation-based methods when tested on other data distributions~\cite{saikia2020optimized}.

\subsubsection{Multi-domain Representations.}
A multi-domain representation (introduced as ``universal representation''
in~\cite{bilen2017universal}) refers to an image representation that works
equally well for a large number
of visual domains. The simplest way to obtain a multi-domain representation is to
train a separate feature extractor for each visual domain and use only the
appropriate one at test time. To reduce the computational
footprint,~\cite{bilen2017universal} investigates if a single CNN can be useful
to perform image classification on very different domains. To achieve this goal,
the authors propose to share most of the parameters between domains during
training and have a small set of parameters that are domain-specific. Such
adaptive feature sharing is implemented using conditional batch
normalization~\cite{batchnorm}, \textit{i.e.} there is a separate set of batch-norm
parameters for every domain. The work of~\cite{rebuffi2018efficient} extends the
idea of domain-specific computations in a single network and proposes universal
parametric network families, which consist of two parts: 1) a CNN feature
extractor with universal parameters shared across all domains, and 2)
domain-specific modules trained on top of universal weights to maximize the
performance on that domain. It has been found
important~\cite{rebuffi2018efficient} to adapt both shallow and deep layers in a
neural network in order to successfully solve multiple visual domains. We use
the method of this paper in our work when training a parametric network family
to produce a multi-domain representation. In contrast, instead of parallel
adapters, in this work, we use much simpler FiLM layers for domain-specific
computations. Importantly, parametric networks families~\cite{rebuffi2018efficient}
and FiLM~\cite{perez2018film} adapters only provide a way to efficiently
compute multi-domain representation; they are not directly useful for few-shot
learning. However, using our SUR strategy on this representation produces
a useful set of features leading to state-of-the-art results in few-shot learning.

\section{Proposed Approach}
We now present our approach for few-shot learning, starting
with preliminaries.

\subsection{Few-Shot Classification with Nearest Centroid Classifier}\label{sec:prelim}
The goal of few-shot classification is to produce a model which, given a new
learning task and a few labeled examples, is able to generalize to
unseen examples for that task. In other words, the model learns from a small
training set $\mathcal{S} = \{(x_i, y_i)\}_{i=1}^{n_S}$, called a
\textit{support set}, and is evaluated on a held-out test set $\mathcal{Q} =
\{(x^*_j, y^*_j)\}_{j=1}^{n_Q}$, called a \textit{query set}. The $(x_i, y_i)$'s
represent image-label pairs while the pair $(\mathcal{S}, \mathcal{Q})$ 
represents the few-shot task.
To fulfill this objective, the problem is addressed in two steps. During the
meta-training stage, a learning algorithm receives a large dataset
$\mathcal{D}_b$, where it must learn a
general feature extractor $f(\cdot)$.
During the meta-testing stage, one is given a target dataset $\mathcal{D}_t$,
used to repeatedly sample few-shot tasks $(\mathcal{S}, \mathcal{Q})$.
Importantly, meta-training ($\mathcal{D}_b$) and meta-testing ($\mathcal{D}_t$)
datasets have no categories in common.

During the meta-testing stage, we use feature representation $f(\cdot)$ to build
a nearest centroid classifier (NCC), similar
to~\cite{mensink2013distance,snell2017prototypical}. Given a support set
$\mathcal{S}$, for each category present in this set, we build a class centroid
by averaging image representations belonging to this category:

\begin{equation}\label{eq:centroid}
  c_j = \frac{1}{|\mathcal{S}_j|} \sum_{i \in \mathcal{S}_j} {f(x_i)}, \quad \mathcal{S}_j = \{k: y_k = j\}, \quad j = 1, ... ,C.
\end{equation}
\\
To classify a sample $x$, we choose a distance function $d(f(x), c_j)$ to be
negative cosine similarity, as
in~\cite{chen19closerfewshot,gidaris2018unsupervised}, and assign the sample to
the closest centroid $c_j$.

\subsection{Method}\label{sec:method}
With the NCC classifier defined above, we may now formally define the concept of multi-domain
representation and the procedure for selecting relevant features.
Typically, the meta-training set is used to train a set of $K$
feature extractors $\{f_i(\cdot)\}_{i=1}^{K}$ that form a multi-domain set of features.
Each feature extractor maps an input image $x$ into a $d$-dimensional
representation $f_i(x) \in \R^{d}$. 
These features should capture different types of semantics and can be obtained in various 
manners, as detailed
in Section~\ref{sec:universal}.

\subsubsection{Parametric Multi-domain Representations.}
One way to transform a multi-domain set of features into a vectorized multi-domain
representation $f(x)$ is, by concatenating all image representations
from this set (with or without $l2$-normalization). As we show in the
experimental section, directly using such $f(x)$ for classification with NCC does not
work well as many irrelevant features for a new task are present in the representation. 
Therefore, we are interested in implementing a selection mechanism.
In order to do so, we define a selection operation $\odot$ as
follows, given a vector ${\lambda}$ in $\R^K$:

\begin{align}\label{eq:selected}
  {\lambda} \odot f(x) &= {\lambda} \cdot
        \begin{bmatrix}
          \hat{f}_{1}(x) \\
          \vdots \\
          \hat{f}_{K}(x)
        \end{bmatrix} = 
        \begin{bmatrix}
          \lambda_1 \cdot \hat{f}_{1}(x) \\
          \vdots \\
          \lambda_K \cdot \hat{f}_{K}(x)
        \end{bmatrix} = f_\lambda(x),
\end{align}
where $\hat{f}_i(x)$ is simply $f_i(x)$ after $\ell_2$-normalization.
We call $f_{\lambda}(x) \in \mathbb{R}^{K\cdot d}$ a parametrized multi-domain
representation, as it contains information from the whole multi-domain set but the
exact representation depends on the selection parameters ${\lambda}$.
Using this mechanism, it is possible
to select various combinations of features from the multi-domain representation by
setting more than one $\lambda_i$ to non-zero values.

\subsubsection{Finding optimal selection parameters.}
Feature selection is performed during meta-testing by optimizing a probabilistic model,
leading to optimal
parameters~${\lambda}$, given a support set $\mathcal{S}=\{(x_i,
y_i)\}_{i=1}^{n_S}$ of a new task.

Specifically, we consider the NCC classifier from Section~\ref{sec:prelim},
using $f_\lambda(x)$ instead of $f(x)$, and introduce the likelihood function
 \begin{equation}\label{eq:probs}
    p(y = l | x) = \frac{\exp(-d(f_\lambda(x), c_l))}{\sum_{j=1}^{n_S}{\exp(-d(f_\lambda(x), c_j))}}.
\end{equation}
Our goal is then to find optimal parameters ${\lambda}$ that
maximize  the likelihood on the support set, which is equivalent to
minimizing the negative log-likelihood:
\begin{align}\label{eq:loss}
  L({\lambda}) &= \frac{1}{n_S} \sum_{i=1}^{n_S}\left[{-\log(p(y = y_i | x_i))} \right] \\
             &= \frac{1}{n_S} \sum_{i=1}^{n_S} \left[ \log \sum_{j=1}^{C} \exp( \cos(f_\lambda(x_i), c_j) ) - \cos (f_\lambda(x_i), c_{y_i}) \right].
\end{align}
This objective is similar to the one of~\cite{wang2018cosface} and encourages
large lambda values to be assigned to representations where intra-class similarity is high
while the inter-class similarity is low. In practice, we optimize the objective
by performing several steps of gradient descent. The proposed procedure is what
we call SUR.

It is worth noting that nearest centroid classifier is a simple non-parametric
model with limited capacity, which only stores a single vector to describe a
class. Such limited capacity becomes an advantage when only a few annotated
samples are available, as it effectively prevents overfitting. When training
and testing across similar domains, SUR is able to select from the multi-domain
representation the features optimized for each visual domain. 
When the target domain distribution does not match any of the train distribution,
it is nevertheless able to adapt a few parameters $\lambda$ to the target distribution.
In such a sense, our method performs a limited form of adaptation, with
few parameters only, which is reasonable given that the target task has only a
small number of annotated samples.

\subsubsection{Sparsity in selection.} As said above, our selection
algorithm is a form of weak adaptation, where parameters ${\lambda}$ are adjusted
to optimize image representation, given an input task. Selection parameters
$\lambda_i$ are constrained to be in $[0, 1]$.
However, as we show in the experimental section, the resulting ${\lambda}$
vector is sparse in practice, with many entries equal to 0 or 1. This empirical behavior
suggests that our algorithm indeed performs selection of relevant features -- a
simple and robust form of adaptation. To promote further sparsity, one may use
an additional sparsity-inducing penalty such as $\ell_1$ during the
optimization~\cite{mairal2014sparse}; however, our experiments show that doing
so is not necessary to achieve good results.

\subsection{Obtaining Multi-Domain Representations}\label{sec:universal}
\begin{figure}[t!]
\centering
  \subfloat[][]{\includegraphics[page=2,width=.40\linewidth,trim=80 25 305 0,clip]{images/Universal_repr.pdf}}
  \subfloat[][]{\includegraphics[page=1,width=.40\linewidth,trim=70 85 275 0,clip]{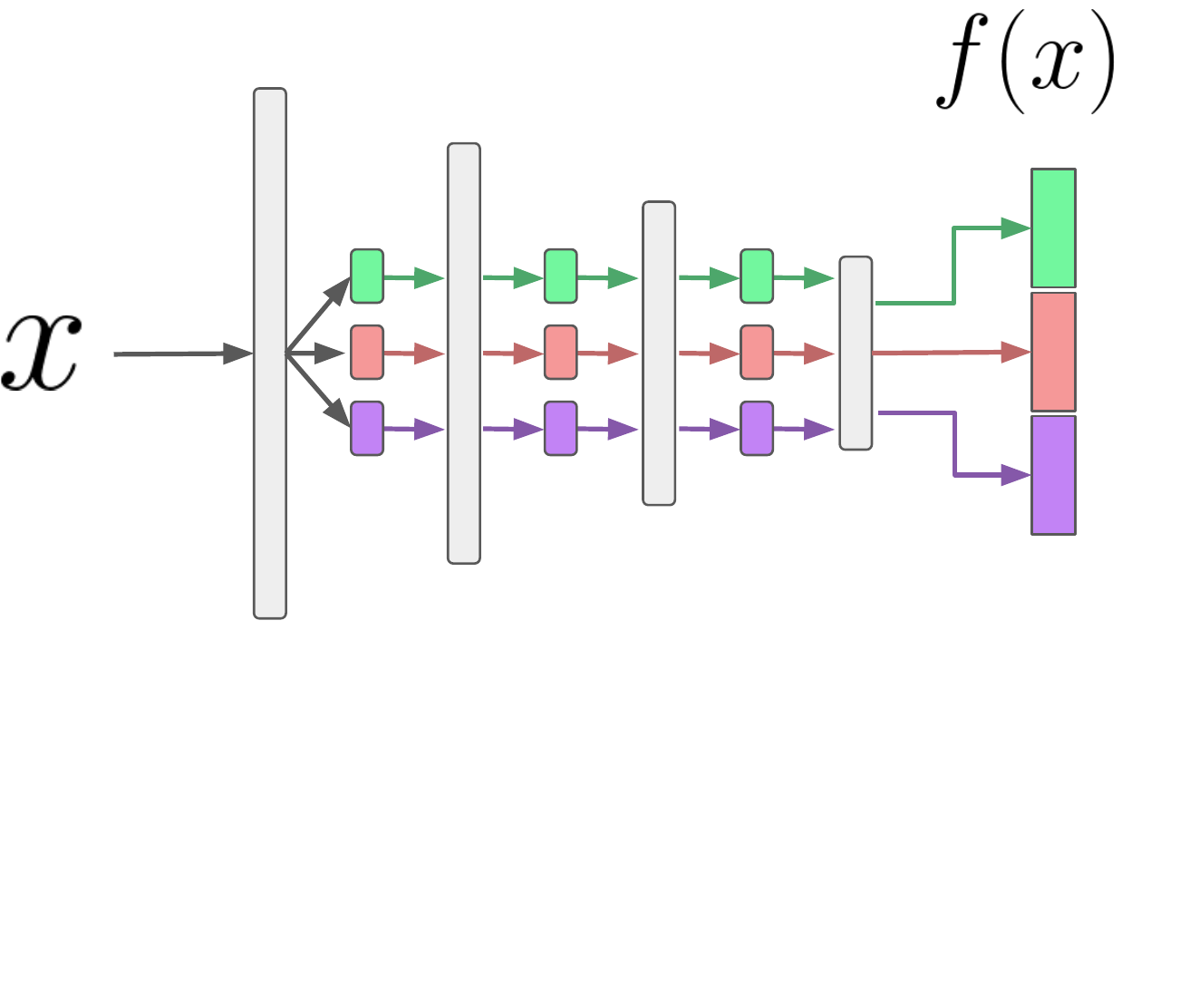}}
\caption{Different ways of obtaining a multi-domain representation. (a) A single
  image is embedded with multiple domain-specific networks. (b) Using a parametric
  network family~\cite{rebuffi2018efficient} to obtain multi-domain
  representations. Here, the gray blocks correspond to
  shared computations and colored blocks correspond to domain-specific
  FiLM~\cite{perez2018film} layers.
  Gray blocks and arrows
  indicate shared layers and computation flow respectively, while domain specific
  ones are shown in corresponding colors. Best viewed in color.}
\label{fig:universal}
\end{figure}

In this section, we explain how to obtain a multi-Domain set of feature
extractors $\{f_i(\cdot)\}_i^{K}$ when one or multiple domains are available for
training. Three variants are used in this paper, which are illustrated in Figure~\ref{fig:universal}.

\subsubsection{Multiple training domains.}\label{subsec:pf}
In the case of multiple training domains, we assume that $K$ different datasets
(including ImageNet~\cite{imagenet})
are available for building a multi-domain representation. We start with a
straightforward solution and train a feature extractor $f_i(\cdot)$
for each visual domain independently. That results in a desired multi-domain set
$\{f_i(\cdot)\}_i^{K}$. A multi-domain representation is then computed by
concatenating the output set, as illustrated in Figure~\ref{fig:universal}(a).

To compute multi-domain representations with a single network, we use parametric
network families proposed in~\cite{rebuffi2018efficient}. We follow the original
paper and first train a general feature extractor -- ResNet -- using our training
split of ImageNet, and then freeze the network's weights. For each of the $K-1$
remaining datasets, task-specific parameters are learned by minimizing the
corresponding training loss. We use FiLM~\cite{perez2018film} layers as
domain-specific modules and insert them after each batch-norm layer in the
ResNet. We choose to use FiLM layers over originally proposed parallel
adapters~\cite{rebuffi2018efficient} because FiLM is much simpler, i.e. performs
channel-wise affine transformation, and contain much fewer parameters. This
helps to avoid overfitting on small datasets.
In summary, each of the $K$ datasets has
its own set of FiLM layers and the base ResNet with a set of domain-specific FiLM
layers constitutes the multi-domain set of extractors $\{f_i(\cdot)\}_i^{K}$. To
compute the features of $i$-th domain $f_i(x)$, we forward the input image $x$
trough the ResNet where all the intermediate activations are modulated with the
FiLM layers trained on this domain, as illustrated in
Figure~\ref{fig:universal}(b). Using parametric network families instead of $K$
separate networks to obtain a multi-domain representation reduces the number of
stored parameters roughly in $K$ times. However, to be actually useful for
few-shot learning, such representation must be processed by our SUR approach, as
described in Section~\ref{sec:method}


\section{Experiments}
We now present the experiments to analyze the performance
of our selection strategy, starting with implementation details.

\subsection{Datasets and Experiments Details}\label{sec:details}
\subsubsection{Datasets.} We use \textit{mini-}ImageNet~\cite{ravioptimization} and 
Meta-Dataset~\cite{triantafillou2019meta} to evaluate the proposed approach. The
\textit{mini}-ImageNet~\cite{ravioptimization} dataset consists of 100 categories (64 for
training, 16 for validation, 20 for testing) from the original
ImageNet~\cite{imagenet} dataset, with 600 images per class. Since all the
categories come from the same dataset, we use \textit{mini}-ImageNet to
evaluate our feature selection strategy in
single-domain few-shot learning. During testing on
\textit{mini}-ImageNet, we measure performance over tasks where only 1 or 5
images (shots) per category are given for adaptation and the number of classes
in a task is fixed to 5, \textit{i.e.} 5-way classification. All images are resized to
$84 \times 84$, as suggested originally by~\cite{ravioptimization}.

Meta-Dataset~\cite{triantafillou2019meta} is much larger than previous few-shot
learning benchmarks and it is actually a collection of multiple
datasets with different data distributions.
It includes 
\texttt{ImageNet}~\cite{imagenet}, \texttt{Omniglot}~\cite{lake2015human},
\texttt{Aircraft}~\cite{maji13fine-grained},
\texttt{CU-Birds}~\cite{wah2011caltech}, Describable
\texttt{Textures}~\cite{cimpoi2014describing}, \texttt{Quick
Draw}~\cite{quickdraw}, \texttt{Fungi}~\cite{fungi},
\texttt{VGG-Flower}~\cite{nilsback2008automated}, \texttt{Traffic
Sign}~\cite{houben2013detection} and \texttt{MSCOCO}~\cite{coco}. A short
description of each dataset is contained in Appendix.
\texttt{Traffic Sign} and \texttt{MSCOCO} datasets are reserved for testing
only, while all other datasets have their corresponding train, val and test
splits. To better study out-of-training-domain behavior, we
follow~\cite{requeima2019fast} and add 3 more testing datasets, namely
\texttt{MNIST}~\cite{mnist} \texttt{CIFAR10}~\cite{krizhevsky2009learning}, and
\texttt{CIFAR100}~\cite{krizhevsky2009learning}. Here, the number of shots and
ways is not fixed and varies from one few-shot task to another. As originally
suggested~\cite{triantafillou2019meta}, all images are
resized to $84 \times 84$ resolution.

\subsubsection{Implementation Details.}
When experimenting with Meta-Dataset, we
follow~\cite{requeima2019fast} and use ResNet18~\cite{resnet} as feature
extractor. The training detail for each dataset are described in Appendix.
To report test results on Meta-Dataset, we perform an independent evaluation
for each of the 10 provided datasets, plus for 3 extra datasets as suggested
by~\cite{requeima2019fast}. We follow~\cite{triantafillou2019meta} and sample
600 tasks for evaluation on each dataset within Meta-Dataset.

When experimenting with \textit{mini}-ImageNet, we follow popular
works~\cite{gidaris2019boosting,lifchitz2019dense,oreshkin2018tadam} and use
ResNet12~\cite{oreshkin2018tadam} as a feature extractor. Corresponding training
details are reported in Appendix. During testing, we use
\textit{mini}-ImageNet's test set to sample 1000 5-way classification tasks. We
evaluate scenarios where only 1 or 5 examples (shots) of each category is
provided for training and 15 for evaluation.
\\
On both datasets, during meta-training, we use cosine classifier with learnable
softmax temperature~\cite{chen19closerfewshot}. During testing, classes and
corresponding train/test examples are sampled at random. For all our
experiments, we report the mean accuracy (in \%) over all test tasks with $95
\%$ confidence interval.

\subsubsection{Feature Selection.} To perform feature selection from the
multi-domain representation we optimize the selection parameter ${\lambda}$
(defined in Eq.~\ref{eq:selected}) to minimize NCC classification loss
(Eq.~\ref{eq:loss}) on the support set. Each individual scalar weight
$\lambda_i$ is kept between 0 and 1 using sigmoid function, \textit{i.e.} $\lambda_i =
\text{sigmoid}(\alpha_i)$. All $\alpha_i$ are initialized with zeros. We
optimize the parameters $\alpha_i$ using gradient descent for 40
iterations. At each iteration, we use the whole support set to build nearest
centroid classifier, and then we use the same set of examples to compute the
loss, given by Eq.~\ref{eq:loss}. Then, we compute gradients w.r.t $[\alpha_1,
..., \alpha_K]$ and use Adadelta~\cite{zeiler2012adadelta} optimizer with
learning rate $10^2$ to perform parameter updates.

\subsection{Cross-Domain Few-shot Classification}\label{sec:cross_domain}
In this section, we evaluate the ability of SUR to handle different visual
domains in MetaDataset~\cite{triantafillou2019meta}. First, we motivate the use
of multi-domain representations and show the importance of feature selection. Then,
we evaluate the proposed strategy against important baselines and
state-of-the-art few-shot algorithms.

\subsubsection{Evaluating domain-specific feature extractors.}
MetaDataset includes 8 datasets for training, \textit{i.e.} \texttt{ImageNet},
\texttt{Omniglot}, \texttt{Aircraft}, \texttt{CU-Birds}, \texttt{Textures},
\texttt{Quick Draw}, \texttt{Fungi}, \texttt{VGG-Flower}. We treat each dataset
as a separate visual domain and obtain a multi-domain set of features by training 8
domain-specific feature extractors, \textit{i.e.} a separate ResNet18 for each dataset.
Each extractor is trained independently, with its own
training schedule specified in Appendix. We test the performance of each feature
extractor (with NCC) on every test dataset specified in
Section~\ref{sec:details}, and report the results in
Table~\ref{tab:feature_results}. Among the 8 datasets seen during training, 5
datasets are better solved with their own features, while 3 other datasets
benefit more from ImageNet features. In general, ImageNet features suits 8
out of 13 test datasets best, while 5 others require a different feature extractor
for better accuracy.
Such results suggest that none of the domain-specific feature extractors alone
can perform equally well on all the datasets simultaneously. However, using the
whole multi-domain feature set to select appropriate representations would lead to
superior accuracy.

\begin{table}[t!] 
\begin{center}
\small\addtolength{\tabcolsep}{-10pt}
\renewcommand{\arraystretch}{1.0}
\renewcommand{\tabcolsep}{1.6mm}
\resizebox{1.0\textwidth}{!}{
\begin{tabular}{l| c c c c c c c c }
\hline
  & \multicolumn{8}{c}{Features trained on:}\\ 
\\[-1em]
 Dataset    & ImageNet   & Omniglot   & Aircraft   & Birds   & Textures  & Quick Draw   & Fungi      & VGG Flower   \\
\hline
ImageNet       & \bb{56.3\pms{1.0}} & 18.5\pms{0.7} & 21.5\pms{0.8} & 23.9\pms{0.8} & 26.1\pms{0.8} & 23.1\pms{0.8} & 31.2\pms{0.8} & 24.3\pms{0.8} \\
Omniglot     & 67.5\pms{1.2} & \bb{92.4\pms{0.5}} & 55.2\pms{1.3} & 59.5\pms{1.3} & 48.4\pms{1.3} & 80.0\pms{0.9} & 59.7\pms{1.2} & 54.2\pms{1.4} \\
Aircraft     & 50.4\pms{0.9} & 17.0\pms{0.5} & \bb{85.4\pms{0.5}} & 30.9\pms{0.7} & 23.9\pms{0.6} & 25.2\pms{0.6} & 33.7\pms{0.8} & 25.1\pms{0.6} \\
Birds        & \bb{71.7\pms{0.8}} & 13.7\pms{0.6} & 18.0\pms{0.7} & 64.7\pms{0.9} & 20.2\pms{0.7} & 17.9\pms{0.6} & 40.7\pms{0.9} & 24.5\pms{0.8} \\
Textures     & \bb{70.2\pms{0.7}} & 30.6\pms{0.6} & 33.1\pms{0.6} & 37.1\pms{0.6} & 57.3\pms{0.7} & 38.5\pms{0.7} & 50.4\pms{0.7} & 45.4\pms{0.8} \\
Quick Draw   & 52.4\pms{1.0} & 50.3\pms{1.0} & 36.0\pms{1.0} & 38.8\pms{1.0} & 35.7\pms{0.9} & \bb{80.7\pms{0.6}} & 35.4\pms{1.0} & 39.4\pms{1.0} \\
Fungi        & 39.1\pms{1.0} & 10.5\pms{0.5} & 14.0\pms{0.6} & 21.2\pms{0.7} & 15.5\pms{0.7} & 13.0\pms{0.6} & \bb{62.7\pms{0.9}} & 22.6\pms{0.8} \\
VGG Flower   & \bb{84.3\pms{0.7}} & 24.8\pms{0.7} & 44.6\pms{0.8} & 57.2\pms{0.8} & 42.3\pms{0.8} & 36.9\pms{0.8} & 76.1\pms{0.8} & 77.1\pms{0.7} \\
\hline
Traffic Sign & \bb{63.1\pms{0.8}} & 44.0\pms{0.9} & 57.7\pms{0.8} & 61.7\pms{0.8} & 55.2\pms{0.8} & 50.2\pms{0.8} & 53.5\pms{0.8} & 57.9\pms{0.8} \\
MSCOCO       & \bb{52.8\pms{1.0}} & 15.1\pms{0.7} & 21.2\pms{0.8} & 22.5\pms{0.8} & 25.8\pms{0.9} & 19.9\pms{0.7} & 29.3\pms{0.9} & 27.3\pms{0.9} \\
MNIST        & 77.2\pms{0.7} & \bb{90.9\pms{0.5}} & 69.5\pms{0.7} & 74.2\pms{0.7} & 55.9\pms{0.8} & 86.2\pms{0.6} & 69.4\pms{0.7} & 66.9\pms{0.7} \\
CIFAR 10     & \bb{66.3\pms{0.8}} & 33.0\pms{0.7} & 37.8\pms{0.7} & 39.3\pms{0.7} & 39.2\pms{0.7} & 36.1\pms{0.7} & 33.6\pms{0.7} & 38.2\pms{0.7} \\
CIFAR 100    & \bb{55.7\pms{1.0}} & 14.9\pms{0.7} & 22.5\pms{0.8} & 25.6\pms{0.8} & 24.1\pms{0.8} & 21.4\pms{0.7} & 22.2\pms{0.8} & 26.5\pms{0.9} \\
\hline
\end{tabular}
}
\end{center}
\caption{\bb{Performance of feature extractors trained with different datasets
  on Meta-Dataset.} The first column indicates the dataset in Meta-Dataset,
  the first row gives the dataset, used to pre-train the feature
  extractor. The body of the table shows feature extractors' accuracy on
  few-shot classification, when applied with NCC. The average accuracy and $95\%$
  confidence intervals computed over 600 few-shot tasks. The numbers in bold
  indicate that a method has the best accuracy per dataset.}
\label{tab:feature_results}
\end{table}

\subsubsection{Evaluating feature selection.} 
\begin{figure}[t!]
\centering
  \subfloat[][]{\includegraphics[page=1,width=.44\linewidth,trim=0 0 0 0,clip]{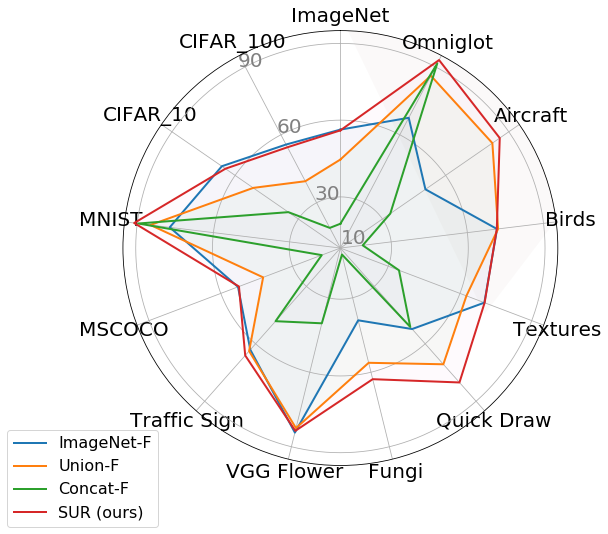}}
  \subfloat[][]{\includegraphics[page=1,width=.44\linewidth,trim=0 0 0 0,clip]{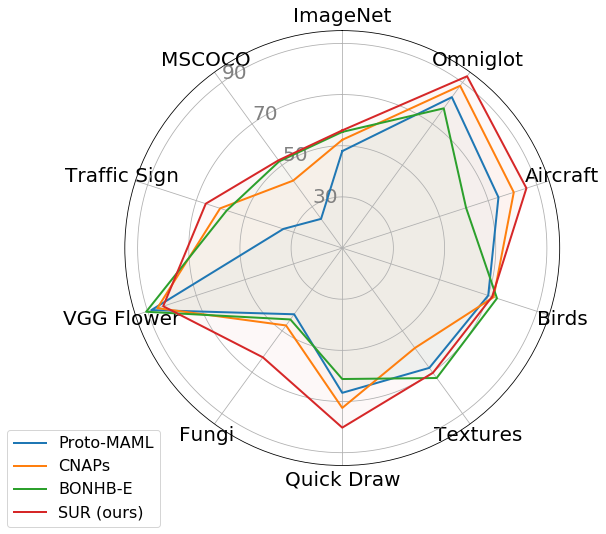}}
\caption{\bb{Performance of different few-shot methods on MetaDataset}.
    (a) Comparison of our selection strategy to baselines. The chart is
    generated from the table in Appendix. (b) Comparing our selection strategy to
    state-of-the-art few-shot learning methods. The chart is generated from
    Table~\ref{tab:metadataset}. The axes indicate the accuracy of methods on a
    particular dataset. Each color corresponds to a different
    method.}
\label{fig:radars}
\end{figure}

We now employ SUR -- our strategy for
feature selection -- as described in Sec.~\ref{sec:method}.
A parametrized multi-domain representation is obtained from a multi-domain feature set
by concatenation. It is then multiplied by the selection parameters
${\lambda}$, that are being optimized for each new few-shot task, following
Section~\ref{sec:details}.
We ran this procedure on all 13 testing datasets and report the results in
Figure~\ref{fig:radars}~(a). We compare our method with the
following baselines: a) using a single feature extractor pre-trained on
\texttt{ImageNet} split of MetaDataset (denoted ``ImageNet-F''), b) using a single feature
extractor pre-trained on the union of 8 training splits in MetaDataset (denoted
``Union-F'') and c)
manually setting all $\lambda_i = 1$, which corresponds to simple concatenation
and (denoted ``Concat-F'').
It is clear that the features provided by SUR have much better
performance than any of the baselines on seen and unseen domains.

\subsubsection{Comparison to other approaches.} We now compare SUR against
state-of-the-art few-shot methods and report the results in
Table~\ref{tab:metadataset}. The results on \texttt{MNIST},
\texttt{CIFAR 10} and \texttt{CIFAR 100} datasets are missing for most of the
approaches because those numbers were not reported in the corresponding original
papers.
Comparison to the best-performing methods on common datasets is summarized
in Figure~\ref{fig:radars}~(b). We see that SUR
demonstrated state-of-the-art results on 9 out of 13 datasets.
BOHNB-E~\cite{saikia2020optimized} outperforms our approach on \texttt{Birds},
\texttt{Textures} and \texttt{VGG Flowers} datasets. This is not surprising
since these are the only datasets that benefit from ImageNet features more than
from their own (see Table~\ref{tab:feature_results}); and
BOHNB-E~\cite{saikia2020optimized} is essentially an ensemble of multiple
ImageNet-pretrained networks.
When tested outside the training domain, SUR consistently outperforms
CNAPs~\cite{requeima2019fast} -- the state-of-the-art adaptation-based method.
Moreover, SUR shows the best results on all 5 datasets never seen during
training.

\begin{table}[t!] 
\begin{center}
\small\addtolength{\tabcolsep}{-10pt}
\renewcommand{\arraystretch}{1.0}
\renewcommand{\tabcolsep}{1.6mm}
\resizebox{1.0\textwidth}{!}{
\begin{tabular}{l| c c c c c c c c}
Test Dataset  &   \makecell{ProtoNet \\ ~\cite{snell2017prototypical}}  &  \makecell{MAML \\ ~\cite{finn2017model}}  &  \makecell{Proto-MAML \\ ~\cite{triantafillou2019meta}}  &  \makecell{CNAPs \\ ~\cite{requeima2019fast}}  &    \makecell{BOHB-E \\ \cite{saikia2020optimized}}     &    \makecell{SUR \\ (ours)}    &    \makecell{SUR-pf \\ (ours)}    &    \makecell{SUR-merge \\ (ours)}   \\
  \hline
ImageNet     &   44.5\pms{1.1} &  32.4\pms{1.0} &  47.9\pms{1.1} &       52.3\pms{1.0} &        55.4\pms{1.1}  &  \bb{56.3\pms{1.1}}  &  \bb{56.4\pms{1.2}} &  \bb{57.2\pms{1.1}}   \\
Omniglot     &   79.6\pms{1.1} &  71.9\pms{1.2} &  82.9\pms{0.9} &       88.4\pms{0.7} &    77.5\pms{1.1}      &  \bb{93.1\pms{0.5}}  &  88.5\pms{0.8}      &  \bb{93.2\pms{0.8}}        \\
Aircraft     &   71.1\pms{0.9} &  52.8\pms{0.9} &  74.2\pms{0.8} &       80.5\pms{0.6} &    60.9\pms{0.9}      &      85.4\pms{0.7}   &  79.5\pms{0.8}      &  \bb{90.1\pms{0.8}}    \   \\
Birds        &   67.0\pms{1.0} &  47.2\pms{1.1} &  70.0\pms{1.0} &       72.2\pms{0.9} &    73.6\pms{0.8}      &  71.4\pms{1.0}       &      76.4\pms{0.9}  &  \bb{82.3\pms{0.8}}        \\
Textures     &   65.2\pms{0.8} &  56.7\pms{0.7} &  67.9\pms{0.8} &       58.3\pms{0.7} &    \bb{72.8\pms{0.7}} &  71.5\pms{0.8}       &  \bb{73.1\pms{0.7}} &  \bb{73.5\pms{0.7}}        \\
Quick Draw   &   65.9\pms{0.9} &  50.5\pms{1.2} &  66.6\pms{0.9} &       72.5\pms{0.8} &    61.2\pms{0.9}      &  \bb{81.3\pms{0.6}}  &  75.7\pms{0.7}      &  \bb{81.9\pms{1.0}}    \   \\
Fungi        &   40.3\pms{1.1} &  21.0\pms{1.0} &  42.0\pms{1.1} &       47.4\pms{1.0} &    44.5\pms{1.1}      &      63.1\pms{1.0}   &  48.2\pms{0.9}      &  \bb{67.9\pms{0.9}}    \   \\
VGG Flower   &   86.9\pms{0.7} &  70.9\pms{1.0} &  88.5\pms{1.0} &       86.0\pms{0.5} &    \bb{90.6\pms{0.6}} &  82.8\pms{0.7}       &  \bb{90.6\pms{0.5}} &      88.4\pms{0.9}        \\
\hline                         
Traffic Sign &   46.5\pms{1.0} &  34.2\pms{1.3} &  34.2\pms{1.3} &       60.2\pms{0.9} &         57.5\pms{1.0} &  \bb{70.4\pms{0.8}}  &  65.1\pms{0.8}      &  67.4\pms{0.8}    \   \\
MSCOCO       &   39.9\pms{1.1} &  24.1\pms{1.1} &  24.1\pms{1.1} &       42.6\pms{1.1} &    \bb{51.9\pms{1.0}} &  \bb{52.4\pms{1.1}}  &  \bb{52.1\pms{1.0}} &      51.3\pms{1.0}    \\
MNIST        &               - &              - &              - &       92.7\pms{0.4} &                     - &  \bb{94.3\pms{0.4}}  &  93.2\pms{0.4}      &  90.8\pms{0.5}        \\
CIFAR 10     &               - &              - &              - &       61.5\pms{0.7} &                     - &  \bb{66.8\pms{0.9}}  &  \bb{66.4\pms{0.8}} &  \bb{66.6\pms{0.8}}   \\
CIFAR 100    &               - &              - &              - &       50.1\pms{1.0} &                     - &      56.6\pms{1.0}   &      57.1\pms{1.0}  &  \bb{58.3\pms{1.0}}   \\

\end{tabular}
}
\end{center}
\caption{\bb{Comparison to existing methods on Meta-Dataset.} The first column
  indicates the of a dataset used for testing. The first row gives a name
  of a few-shot algorithm. The body of the table contains average accuracy and
  $95\%$ confidence intervals computed over 600 few-shot tasks.  The numbers in
  bold have intersecting confidence intervals with the most accurate method.}
\label{tab:metadataset}
\end{table}

\subsubsection{Multi-domain representations with parametric network family.}
While it is clear that SUR outperforms other approaches, one may raise a
concern that the improvement is due to the increased number of parameters, that is,
we use 8 times more parameters than in a single ResNet18.
To address this concern, we use a parametric network
family~\cite{rebuffi2018efficient} that has only $0.5\%$ more parameters than a
single ResNet18. As described in Section~\ref{subsec:pf}, the parametric network
family uses ResNet18 as a base feature extractor and FiLM~\cite{perez2018film}
layers for feature modulation. The total number of additional parameters,
represented by all domain-specific FiLM layers is approximately $0.5\%$ of
ResNet18 parameters. For comparison, CNAPs adaptation mechanism is larger than
ResNet18 itself. To train the parametric network family, we first train a base
CNN feature extractor on ImageNet. Then, for each remaining training dataset, we
learn a set of FiLM layers, as detailed in Section~\ref{subsec:pf}. To obtain a
multi-domain feature set for an image, we run inference 8 times, each time with a
set of FiLM layers, corresponding to a different domain, as described
in~\ref{subsec:pf}. Once the multi-domain feature is built, our selection mechanism
is applied to it as described before (see Section~\ref{sec:details}). The
results of using SUR with a parametric network family are presented in
Table~\ref{tab:metadataset} as ``SUR-pf''. The table suggests that the accuracy
on datasets similar to ImageNet is improved suggesting that parameter
sharing is beneficial in this case and confirms the original findings
of~\cite{rebuffi2018efficient}. However, the opposite is true for different
visual domains such as \texttt{Fungi} and \texttt{QuickDraw}. It implies that to
do well on very different domains, the base CNN filters must be
learned on those datasets from scratch, and feature modulation is not
competitive.

\subsubsection{Reducing the number of training domains.}
Another way to increase SUR's efficiency is to use fewer domains. This means
training fewer domain-specific feature extractors, faster inference and
selection. To achieve this goal, we merge similar datasets together into
separate visual domains, and train a single feature extractor for each such
domain. Here, we take 8 datasets within MetaDataset and use them to form 3
following visual domains: [\texttt{ImageNet}, \texttt{Aircraft},
  \texttt{CU-Birds}, \texttt{Textures}, \texttt{VGG-Flower}],
[\texttt{Omniglot}, \texttt{Quick Draw}], [\texttt{Fungi}, \texttt{ImageNet},
  \texttt{Aircraft}]. Then, we train 3 feature extractors, one for each such
domain. These 3 feature extractors constitute the multi-domain feature set that
is then used for selection with SUR, as described in Section~\ref{sec:details}.
Doing so leads to a 2x speed-up in training and inference. Moreover, as Table~\ref{tab:metadataset} suggests, using
`SUR-merge` allows us to achieve better performance. We attribute
this to learning better representation thanks to the increased number of
relevant training samples per feature extractor.

\begin{figure}[t!]
\begin{center}
  \includegraphics[width=0.99\linewidth]{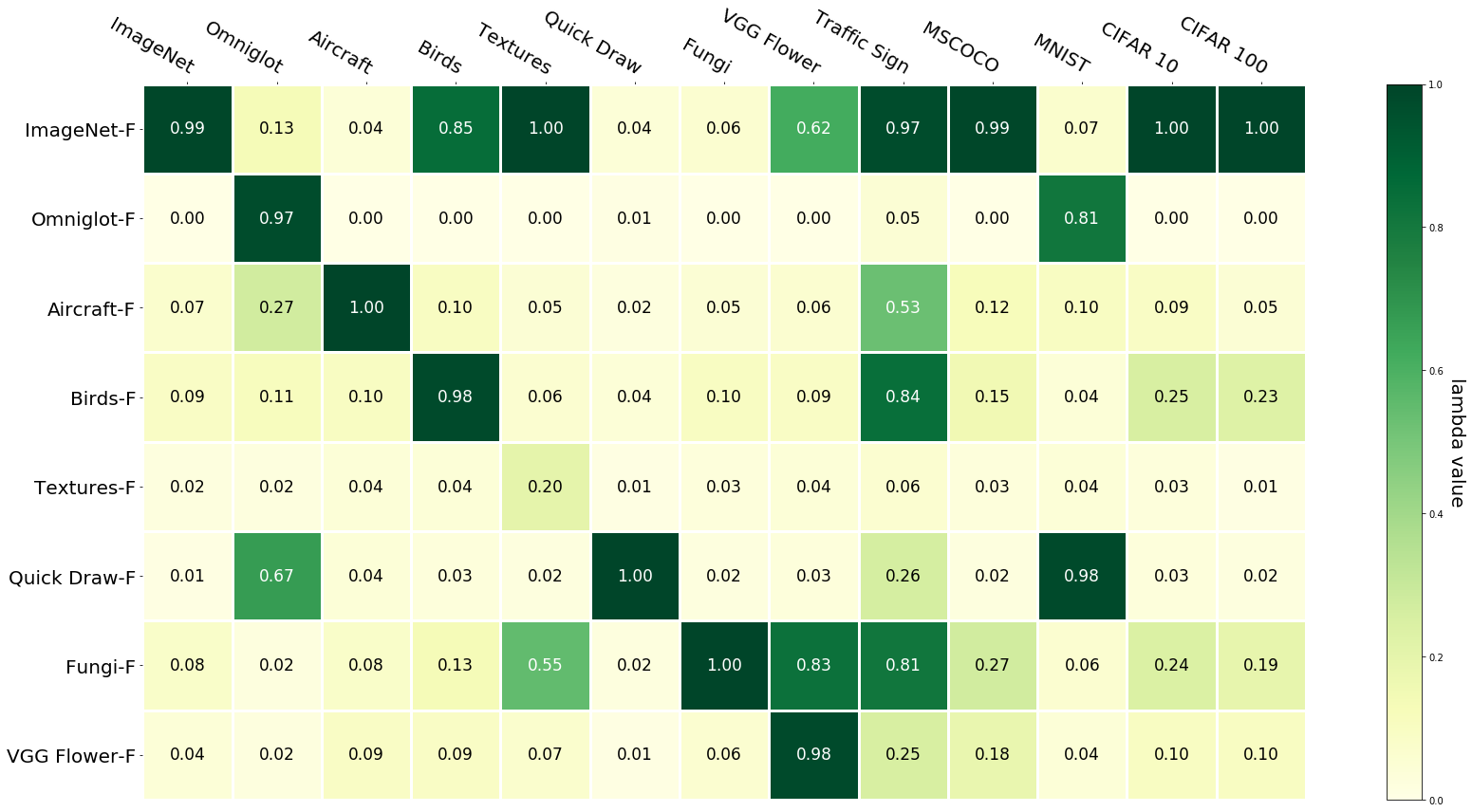}
\end{center}
\caption{\textbf{Frequency of selected features depending on the test domain in MetaDataset.}
  The top row indicates a testing dataset. The leftmost column a dataset the
  feature extractor has been trained on. A cells at location $i, j$ reflects
  the average value of selection parameter $\lambda_i$ assigned to the $i$-th feature
  extractor when tested on $j$-th dataset. The values are averaged over 600
  few-shot test tasks for each dataset.}
\label{fig:lambdas}
\end{figure}

\subsection{Analysis of Feature Selection}
In this section, we analyze optimized selection parameters ${\lambda}$ when
applying SUR on MetaDataset. Specifically, we perform the experiments from
Section~\ref{sec:cross_domain}, where we select appropriate representations from
a set, generated by 8 independent networks. For each test dataset, we then
average selection vectors ${\lambda}$ (after being optimized for 40 SGD
steps) over 600 test tasks and present them in Figure~\ref{fig:lambdas}.
First, we can see that the resulting ${\lambda}$ are sparse, confirming that most of the time
SUR actually select a few relevant features rather than takes all features with
similar weights. Second, for a given test domain, SUR tends to select feature
extractors trained on similar visual domains. Interestingly, for datasets coming
from exactly the same distribution, \textit{i.e.} \texttt{CIFAR 10} and \texttt{CIFAR
  100}, the averaged selection parameters are almost identical. All of the above
suggests that the selection parameters could be interpreted as encoding the
importance of features’ visual domains for the test domain.

\begin{table}[t!] 
\centering
\renewcommand{\arraystretch}{1.0}
\renewcommand{\tabcolsep}{1.6mm}
\resizebox{0.60\linewidth}{!}{
\begin{tabular}{l| c | c c c}
  Method &  Aggregation & 5-shot & 1-shot \\
  \hline
  \multirow{3}{*}{Cls}  & last   & 76.28 \pms{0.41} & 60.09 \pms{0.61} \\
                        & concat & 75.67 \pms{0.41} & 57.15 \pms{0.61} \\
                        & SUR    & {\bf 79.25} \pms{0.41} & {\bf 60.79} \pms{0.62} \\
  \hline
  \hline
  \multirow{3}{*}{DenseCls}  & last   & 78.25 \pms{0.43} & 62.61 \pms{0.61} \\
                             & concat & 79.59 \pms{0.42} & 62.74 \pms{0.61} \\
                             & SUR    & {\bf 80.04} \pms{0.41} & {\bf 63.13} \pms{0.62} \\
  \hline
  \hline
  \multirow{3}{*}{Robust20-dist}  & last   & 81.06 \pms{0.41} & {\bf 64.14} \pms{0.62} \\
                                  & concat    & 80.79 \pms{0.41} & 63.22 \pms{0.63} \\
                                  & SUR & {\bf 81.19} \pms{0.41} & 63.93 \pms{0.63} \\
  \hline
\end{tabular}
}
\caption{\textbf{Comparison to other methods on 1- and 5-shot \textit{mini}-ImageNet.}
  The first column specifies a way of training a feature extractor, while the
  second column reflects how the final image representation is constructed.
  The two last columns display the accuracy on 1- and 5-shot learning tasks. The
  average is reported over $1\,000$ independent experiments
  with $95\%$ confidence interval. The best accuracy is in bold.}
\label{tab:mini_imagenet}
\end{table}

\subsection{Single-domain Few-shot Classification}
In this section, we show that our feature selection strategy can be effective
for various problems, not just for multi-domain few-shot learning. As an
example, we demonstrate the benefits of SUR on few-shot classification,
when training and testing classes come from the same dataset. Specifically,
we show how to use our selection strategy in order to improve existing
adaptation-free methods.
\\
To test SUR in the single-domain scenario we use
\textit{mini}-ImageNet benchmark and solve 1-shot and 5-shot classification, as
described in Section~\ref{sec:details}.
When only one domain is available, obtaining a truly multi-domain set of features is not possible.
Instead, we construct a proxy for such a set by using
activations of network's intermediate layers. In this approximation, each intermediate
layer provides features corresponding to some domain. Since different layers
extract different features, selecting the relevant features should help on new
tasks.
\\
We experiment with 3 adaptation-free methods. They all use the last layer of
ResNet12 as image features and build a NCC on top, however, they differ in a way the
feature extractor is trained. The method we call ``Cls'', simply trains ResNet12
for classification on the meta-training set. The work
of~\cite{lifchitz2019dense} performs dense classification instead (dubbed
``DenseCls''). Finally, the ``Robust20-dist'' feature
extractor~\cite{dvornik2019diversity} is obtained by ensemble distillation.
For any method, the ``multi-domain'' feature set is formed from activations
of the last 6 layers of the network. This is because the remaining intermediate
layers do not contain useful for the final task information, as we show in
Appendix.
\\
Here, we explore different ways of exploiting such ``multi-domain'' set of
features for few-shot classification and report the results in
Table~\ref{tab:mini_imagenet}. We can see that using SUR to select appropriate
for classification layers usually works better than using only
the penultimate layer (dubbed ``last'') or concatenating all the features
together (denoted as ``concat''). For Robust20-dist, we observe only incremental
improvements for 5-shot classification and negative improvement in 1-shot
scenario. The penultimate layer in this network
is probably the most useful for new problems and, if not selected, may hinder
the final accuracy.



\section{Acknowledgments}
This work was funded in part by the French government under management
of Agence Nationale de la Recherche as part of the ``Investissements davenir''
program, reference ANR-19-P3IA-0001 (PRAIRIE 3IA Institute) and reference
ANR-19-P3IA-0003 (3IA MIAI@Grenoble Alpes), and was supported by the ERC grant
number 714381 (SOLARIS) and a gift from Intel.

%
%
\bibliographystyle{splncs04}

\clearpage

\appendix
\section{Implementation and Datasets details}
\subsection{Full MetaDataset Description}
\begin{figure}[th!]
   \centering
  \includegraphics[width=0.57\linewidth,trim=0 0 0 0,clip]{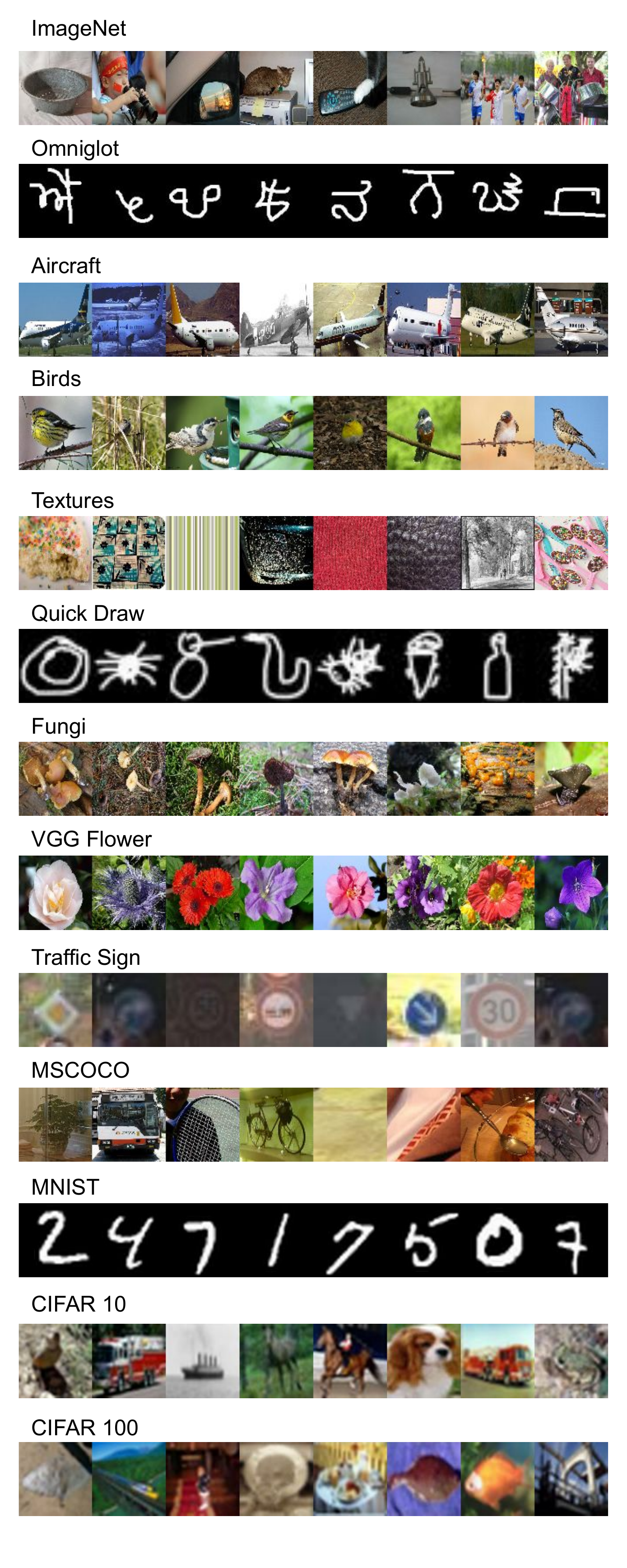}
\caption{\textbf{Samples from all MetaDataset datasets} Each line gives 8 random
samples from a dataset specified above.}\label{fig:datasets}
\end{figure}

The MetaDataset includes
\texttt{ImageNet}~\cite{imagenet} (1000 categories of natural images),
\texttt{Omniglot}~\cite{lake2015human} (1623 categories of black-and-white
hand-written characters from different alphabets),
\texttt{Aircraft}~\cite{maji13fine-grained} (100 classes of aircraft types),
\texttt{CU-Birds}~\cite{wah2011caltech} (200 different bird species),
Describable \texttt{Textures}~\cite{cimpoi2014describing} (43 categories for
textures), \texttt{Quick Draw}~\cite{quickdraw} (345 different categories of
black-and-white sketches), \texttt{Fungi}~\cite{fungi} (1500 mushroom
types), \texttt{VGG-Flower}~\cite{nilsback2008automated} (102
flower species), \texttt{Traffic Sign}~\cite{houben2013detection} (43 classes of
traffic signs) and \texttt{MSCOCO}~\cite{coco} (80 categories of day-to-day
objects). For testing, we additionally employ \texttt{MNIST}~\cite{mnist} (10
hand-written digits)
\texttt{CIFAR10}~\cite{krizhevsky2009learning} (10 classes of common objects), and
\texttt{CIFAR100}~\cite{krizhevsky2009learning} (100 classes of common objects).
Figure~\ref{fig:datasets} illustrated random samples drawn from each dataset.

\subsection{MetaDataset training details}
When using multiple ResNet18 on MetaDataset (a single ResNet per dataset) to
build a multi-domain representation, we train the networks according to the
following procedure.
For optimization, we use SGD with momentum and adjust the learning rate using
cosine annealing~\cite{loshchilov2016sgdr}. The starting learning rate, the maximum number of
training iterations (``Max iter.'') and annealing frequency (``annealing
freq.'') are set individually for each dataset. To regularize training, we use
data augmentation, such as random crops and random color augmentations, and set
a constant weight decay of $7 \times 10^{-4}$. For each dataset, we run a grid
search over batch size in [8, 16, 32, 64] and pick the one that maximizes
accuracy on the validation set. The hyper-parameters maximizing the validation
accuracy are given in Table~\ref{tab:metadataset_hyper}.

When training a parametric network family for building multi-domain
representations, we start by adopting a ResNet18 already trained on ImageNet,
that we keep fixed for the rest of the training procedure. For each new dataset,
we then train a set of domain-specific FiLM layers, modulating intermediate
ResNet layers, as described in Section 3.3
of the original paper. Here, we also use cosine annealing as learning rate
policy, employ weight decay and data augmentation, as specified above. In
Table~\ref{tab:metadataset_pf_hyper}, we report the training hyper-parameters
for each of the datasets.

\subsection{mini-ImageNet training details}
All the methods we evaluate on \textit{mini}-ImageNet use
ResNet12~\cite{oreshkin2018tadam} as a feature extractor. It is trained with
batch size 200 for 48 epochs. For optimization, we use Adam
optimizer~\cite{adam} with initial learning rate 0.1 which is kept constant for
the first 36 epochs. Between epochs 36 and 48, the learning rate was
exponentially decreased from 0.1 to $10^{-5}$, \textit{i.e.} by dividing the learning rate by
$10^{\frac{1}{3}}$ after each epoch. As regularization, we use weight decay with
$5\times 10^{-4}$ multiplier and data augmentation such as random crops, flips
and color transformations.

\begin{table}[t!] 
\begin{center}
\small\addtolength{\tabcolsep}{-10pt}
\renewcommand{\arraystretch}{1.0}
\renewcommand{\tabcolsep}{1.6mm}
\resizebox{\textwidth}{!}{
\begin{tabular}{l| c c c c c}
Test Dataset &   learning rate   &    weight decay   &    Max iter.  & annealing freq.  & batch size \\
\hline
ImageNet     & $3 \times 10^{-2}$ &  $7 \times 10^{-4}$&       480,000 &      48,000 &      64   \\
Omniglot     & $3 \times 10^{-2}$ &  $7 \times 10^{-4}$&       50,000  &      3,000  &      16   \\
Aircraft     & $3 \times 10^{-2}$ &  $7 \times 10^{-4}$&       50,000  &      3,000  &      8    \\
Birds        & $3 \times 10^{-2}$ &  $7 \times 10^{-4}$&       50,000  &      3,000  &      16   \\
Textures     & $3 \times 10^{-2}$ &  $7 \times 10^{-4}$&       50,000  &      1,500  &      32   \\
Quick Draw   & $1 \times 10^{-2}$ &  $7 \times 10^{-4}$&       480,000 &      48,000 &      64   \\
Fungi        & $3 \times 10^{-2}$ &  $7 \times 10^{-4}$&       480,000 &      15,000 &      32   \\
VGG Flower   & $3 \times 10^{-2}$ &  $7 \times 10^{-4}$&       50,000  &      1,500  &      8    \\
\hline 
\end{tabular}
}
\end{center}
\caption{\bb{Training hyper-parameters of individual feature networks on MetaDataset}. The first column
  indicates the dataset used for training. The first row gives the name of he
  hyper-parameter. The body of the table contains hyper-parameters that produced
  the most accurate model on the validation set.}
\label{tab:metadataset_hyper}
\end{table}

\begin{table}[t!] 
\begin{center}
\small\addtolength{\tabcolsep}{-10pt}
\renewcommand{\arraystretch}{1.0}
\renewcommand{\tabcolsep}{1.6mm}
\resizebox{\textwidth}{!}{
\begin{tabular}{l| c c c c c}
Test Dataset &   learning rate   &    weight decay   &    Max iter.  & annealing freq.  & batch size \\
\hline
Omniglot     & $3 \times 10^{-2}$ &  $7 \times 10^{-4}$&       40,000  &      3,000  &      16   \\
Aircraft     & $1 \times 10^{-2}$ &  $7 \times 10^{-4}$&       30,000  &      1,500  &      32    \\
Birds        & $3 \times 10^{-2}$ &  $7 \times 10^{-4}$&       30,000  &      1,500  &      16   \\
Textures     & $3 \times 10^{-2}$ &  $7 \times 10^{-4}$&       40,000  &      1,500  &      16   \\
Quick Draw   & $1 \times 10^{-2}$ &  $7 \times 10^{-4}$&       400,000 &      15,000 &      32   \\
Fungi        & $1 \times 10^{-2}$ &  $7 \times 10^{-4}$&       400,000 &      15,000 &      32   \\
VGG Flower   & $1 \times 10^{-2}$ &  $7 \times 10^{-4}$&       30,000  &      3,000  &      16    \\
\hline 
\end{tabular}
}
\end{center}
\caption{\bb{Training hyper-parameters of the parametric network family on
    MetaDataset}. The first column indicates the dataset used for training. The
  first row gives the name of he hyper-parameter. The body of the table contains
  hyper-parameters that produced the most accurate model on the validation set.}
\label{tab:metadataset_pf_hyper}
\end{table}

\section{Additional Experiments and Ablation Study}
\subsection{Additional results on MetaDataset}
Here we elaborate on using SUR with a multi-domain set of representations obtained from independent
feature extractors (see Section 3.2), report an ablation study on varying the
number of extractors in the multi-domain set, and report detailed results,
corresponding to Figure 3 (a) of the original paper. Specifically, 
we use 8 domain-specific ResNet18 feature extractors to build a multi-domain
representation and evaluate SUR against the baselines. The results are reported
in Table~\ref{tab:baseline}, which corresponds to Figure 3 (a) of the original
paper.

In the following experiment, we remove feature extractors trained on
\texttt{Birds}, \texttt{Textures} and \texttt{VGG Flower} from the multi-domain
feature set and test the performance of SUR on the set of remaining 5 feature
extractors. We chose to remove these feature extractors as none of them gives
the best performance on any of the test sets. Hence, they probably do not add new
knowledge to the multi-domain set of features. The results are reported in
Table~\ref{tab:baseline}~(a) as ``SUR (5/8)''. As we can see, selecting from the
truncated set of features may be beneficial for some out-of-domain categories,
which suggests that even the samples form of adaptation -- selection -- may
overfit when very few samples are available. On the other hand, for a new
dataset \texttt{Traffic Sign}, selecting from all features is beneficial.
This result is not surprising, as one generally does not know what features will
be useful for tasks not known beforehand, and thus removing seemingly useless
features may result in a performance drop.

\begin{table}[t!] 
\begin{center}
\small\addtolength{\tabcolsep}{-10pt}
\renewcommand{\arraystretch}{1.0}
\renewcommand{\tabcolsep}{1.6mm}
\resizebox{0.78\textwidth}{!}{
\begin{tabular}{l| c c c c c}
Test Dataset &   ImageNet-F           &    Union-F   &    Concat-F  & SUR  & SUR (5/8) \\
\hline
ImageNet     &   \bb{56.3\pms{1.0}} &  44.6\pms{0.7}&       19.5\pms{1.6} &      \bb{56.0\pms{1.2}}  &      \bb{56.1\pms{1.1}}   \\
Omniglot     &   67.5\pms{1.2}      &  86.1\pms{0.9}&       91.5\pms{0.5} &      \bb{93.0\pms{0.4}}  &      \bb{93.1\pms{0.4}}   \\
Aircraft     &   50.4\pms{0.9}      &  82.2\pms{0.6}&       33.7\pms{1.4} &      \bb{85.7\pms{0.3}}  &      \bb{85.5\pms{0.4}}   \\
Birds        &   71.7\pms{0.8}      &  \bb{72.1\pms{1.1}}&  18.8\pms{1.3} &      71.6\pms{0.8}       &      71.0\pms{0.8}        \\
Textures     &   \bb{70.2\pms{0.7}} &  62.7\pms{1.0}&       34.5\pms{0.9} &      \bb{70.3\pms{0.9}}  &      \bb{70.4\pms{0.9}}   \\
Quick Draw   &   52.3\pms{1.0}      &  70.7\pms{0.9}&       51.2\pms{0.9} &      \bb{80.2\pms{0.8}}  &      \bb{80.5\pms{0.9}}   \\
Fungi        &   39.1\pms{1.0}      &  56.2\pms{0.8}&       12.6\pms{0.4} &      \bb{62.8\pms{1.1}}  &      \bb{63.1\pms{1.0}}   \\
VGG Flower   &   \bb{84.3\pms{0.7}} &  82.5\pms{0.8}&       40.3\pms{1.2} &      83.6\pms{0.8}       &      83.3\pms{0.8}        \\
\hline                             
Traffic Sign & 63.1\pms{0.8}        & 63.8\pms{0.9}  &    48.2\pms{0.6}   &       \bb{66.1\pms{0.8}} &       63.6\pms{0.9}   \\
MSCOCO       & \bb{52.8\pms{1.0}}   & 42.3\pms{1.0}  &    17.8\pms{0.4}   &       \bb{52.4\pms{1.1}} &       \bb{52.8\pms{1.1}}   \\
MNIST        & 77.2\pms{0.7}        & 84.8\pms{0.6}  &    89.6\pms{0.7}   &       91.2\pms{0.5}      &       \bb{92.5\pms{0.5}}   \\
CIFAR 10     & \bb{66.3\pms{0.8}}   & 51.4\pms{0.8}  &    34.7\pms{0.8}   &       64.6\pms{0.9}      &       65.8\pms{0.9}        \\
CIFAR 100    & \bb{55.7\pms{1.0}}   & 39.5\pms{1.0}  &    18.9\pms{0.6}   &       54.5\pms{1.0}      &       \bb{56.5\pms{1.0}}        \\

\end{tabular}
}
\end{center}
\caption{\bb{Motivation for feature selection.} The table shows accuracy of different
  feature combinations on the Meta-Detaset test splits. The first column
  indicates the dataset the algorithms are tested on, the first row gives a name
  of a few-shot algorithm. The body of the table contains average accuracy and
  $95\%$ confidence intervals computed over 600 few-shot tasks. The numbers in
  bold lie have intersecting confidence intervals with the most accurate method.}
\label{tab:baseline}
\end{table}

\begin{table}[t!] 
\centering
\renewcommand{\arraystretch}{1.0}
\renewcommand{\tabcolsep}{1.6mm}
\resizebox{0.79\linewidth}{!}{
\begin{tabular}{l| c c c c c | c c c}
  Method & 1-3 & 4-6 & 7-9 & 10-12 & Aggregation & 5-shot & 1-shot \\
  \hline
  \multirow{5}{*}{Cls}         & & & &   & last & 76.28 \pms{0.41} & 60.09 \pms{0.61} \\
           &            &            &            &  \checkmark  &select & 77.39 \pms{0.42} & \bf{ 61.02} \pms{0.62} \\
           &            &            & \checkmark &  \checkmark  &select& {\bf 79.25} \pms{0.41} & 60.79 \pms{0.62} \\
           &            & \checkmark & \checkmark &  \checkmark  &select & 78.92 \pms{0.41} & 60.71 \pms{0.64} \\
           & \checkmark & \checkmark & \checkmark &  \checkmark  &select &  78.80 \pms{0.43} & 60.55 \pms{0.62} \\

           &  & & & \checkmark  & concat & 78.43 \pms{0.42} & 60.41 \pms{0.62} \\
           &  &  & \checkmark& \checkmark  & concat & 75.67 \pms{0.41} & 57.15 \pms{0.61} \\
           &  & \checkmark & \checkmark &\checkmark &concat & 70.90 \pms{0.40} & 53.53 \pms{0.61} \\
           & \checkmark & \checkmark & \checkmark&  \checkmark  &concat &  69.40 \pms{0.40} & 51.21 \pms{0.60} \\
  \hline
  \hline
  \multirow{5}{*}{DenseCls}         &  &  &  &  & last& 78.25 \pms{0.43} & 62.61 \pms{0.61} \\
           &            &            &            &  \checkmark & select& 79.34 \pms{0.42} & 62.46 \pms{0.62} \\
           &            &            & \checkmark &  \checkmark & select & {\bf 80.04} \pms{0.41} & {\bf 63.13} \pms{0.62} \\
           &            & \checkmark & \checkmark &  \checkmark & select & 79.84 \pms{0.42} & 62.95 \pms{0.62} \\
           & \checkmark & \checkmark & \checkmark &  \checkmark & select& 79.49 \pms{0.43} & 62.58 \pms{0.63} \\

           &            &            &            &  \checkmark & concat& 79.12 \pms{0.41} & 62.51 \pms{0.62} \\
           &            &            & \checkmark &  \checkmark & concat & 79.59 \pms{0.42} & 62.74 \pms{0.61} \\
           &            & \checkmark & \checkmark &  \checkmark & concat & 77.63 \pms{0.42} & 60.14 \pms{0.61} \\
           & \checkmark & \checkmark & \checkmark &  \checkmark & concat & 76.07 \pms{0.41} & 57.78 \pms{0.61} \\
  \hline
  \hline
  \multirow{5}{*}{DivCoop}         &  &  &  &   & last & 81.06 \pms{0.41} & \bf{64.14} \pms{0.62} \\
           &            &            &            &  \checkmark & select& {\bf 81.23} \pms{0.42} & 63.83 \pms{0.62} \\
           &            &            & \checkmark &  \checkmark & select   & 81.19 \pms{0.41} & 63.93 \pms{0.63} \\
           &            & \checkmark & \checkmark &  \checkmark & select& 81.11 \pms{0.42} & 63.85 \pms{0.62} \\
           & \checkmark & \checkmark & \checkmark &  \checkmark & select&  81.08 \pms{0.42} & 63.71 \pms{0.62} \\

           &            &            &            &  \checkmark & concat& 81.12 \pms{0.42} & 63.92 \pms{0.62} \\
           &            &            & \checkmark &  \checkmark & concat   & 80.79 \pms{0.41} & 63.22 \pms{0.63} \\
           &            & \checkmark & \checkmark &  \checkmark & concat& 80.52 \pms{0.42} & 62.48 \pms{0.61} \\
           & \checkmark & \checkmark & \checkmark &  \checkmark & concat&  80.36 \pms{0.42} & 61.30 \pms{0.61} \\
  \hline
\end{tabular}
}
\caption{\textbf{Comparison to other methods on 1- and 5-shot \textit{mini}-ImageNet.}
  The first column gives the name of the feature extractor. Columns 2-5 indicate
  if corresponding layers of ResNet12 were added to the multi-domain set of
  representations. Column ``Aggregation'' specifies how the multi-domain set was
  used to obtain a vector image representation.
  The two last columns display the accuracy on 1- and 5-shot learning tasks. To
  evaluate our methods we performed $1\,000$ independent experiments on
  \textit{mini}-ImageNet-test and report the average and $95\%$ confidence
  interval. The best accuracy is in bold.}
\label{tab:mini_imagenet_all}
\end{table}

\subsection{Analysis of Feature Selection on MetaDataset}
\begin{figure}[t!]
\begin{center}
  \includegraphics[width=0.99\linewidth]{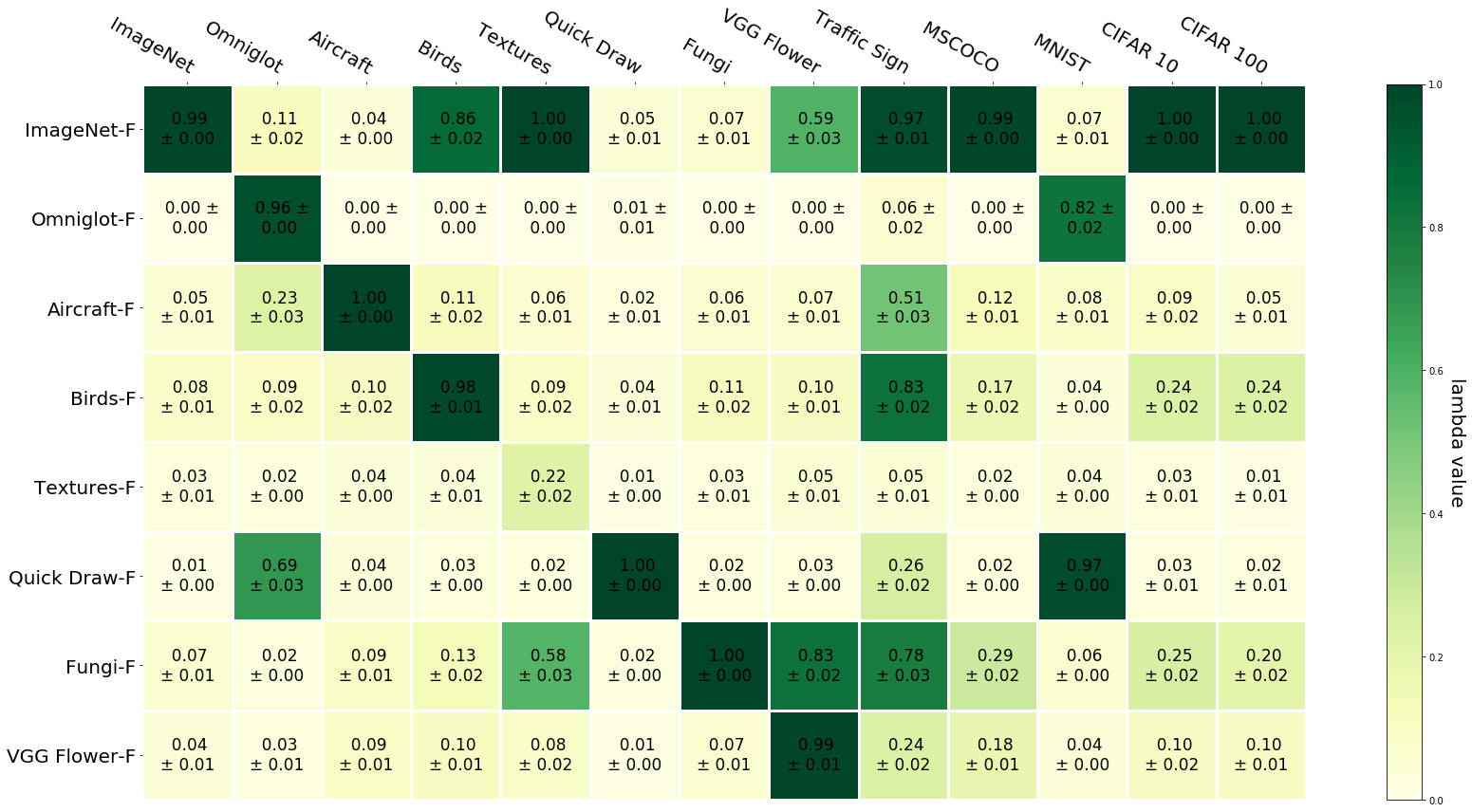}
\end{center}
\caption{\textbf{Frequency of selected features depending on the test domain in MetaDataset.}
  The top row indicates a testing dataset. The leftmost column presents a dataset the
  feature extractor has been trained on. A cells at location $i, j$ reflects
  the average value of selection parameter $\lambda_i$ assigned to the $i$-th feature
  extractor when tested on $j$-th dataset with corresponding $95\%$ confidence
  intervals. The values are averaged over 600 few-shot test tasks for each
  dataset.}
\label{fig:lambdas_conf}
\end{figure}

Here, we repeat the experiment from Section 4.3, \textit{i.e.} studying average values of
selection parameters $\lambda$ depending on the test dataset.
Figure~\ref{fig:lambdas_conf} reports the average selection parameters with
corresponding confidence intervals. This is in contrast to Figure 4 of the
original paper that reports the average values only, without confidence intervals.

\begin{figure}[t!]
\begin{center}
  \includegraphics[width=0.99\linewidth]{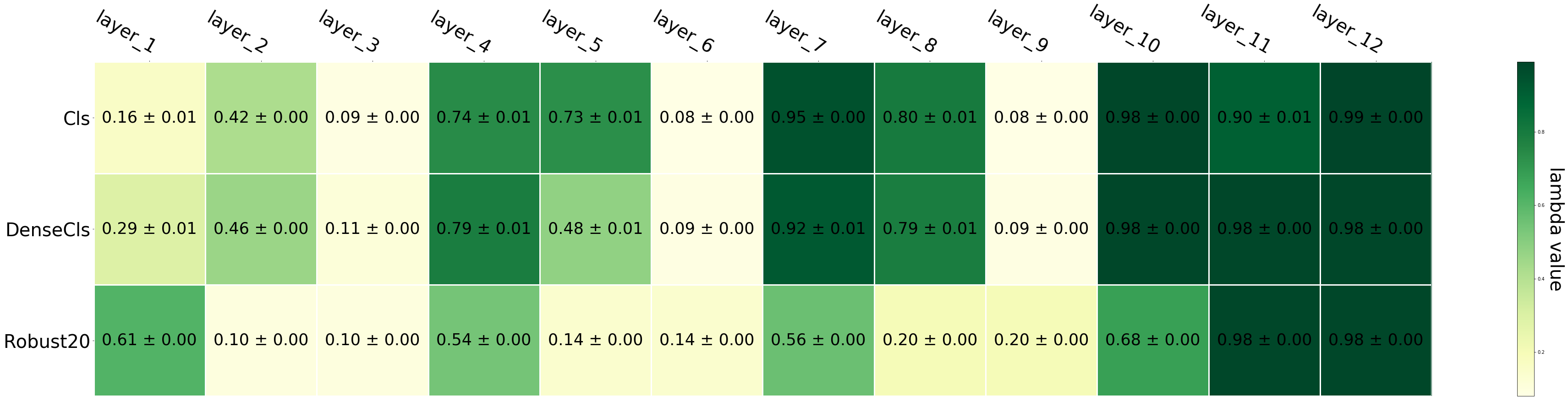}
\end{center}
\caption{\textbf{Frequency of selecting intermediate layer's activation's on
    mini-ImageNet for 5-shot classification.}
  The top row indicates intemediate layer. The leftmost column gives the name of
  a method used to pre-train the feature extractor. Each cells reflects
  the average value of selection parameter $\lambda_i$ assigned to the $i$-th
  intemediate layer with corresponding $95\%$ confidence intervals. The values
  are averaged over 1000 few-shot test tasks for each dataset.}
\label{fig:lambdas_mini_imagenet}
\end{figure}

\subsection{Importance of Intermediate Layers on mini-ImageNet}
We clarify the findings in Section~4.4 of the original paper
and provide an ablation study on the importance of intermediate layers
activations for the meta-testing performance. For all experiments on \textit{mini}-ImageNet, we
use ResNet12 as a feature extractor and construct a multi-domain feature set from
activations of intermediate layers. In Table~\ref{tab:mini_imagenet_all}, we
experiment with adding different layers outputs to the multi-domain set. The
multi-domain set is then used to
construct the final image representation either through concatenation ``concat''
or using SUR. The table suggests that adding the first 6 layers negatively
influences the performance of the target task. While our SUR approach can still
select relevant features from the full set of layers, the negative impact is
especially pronounced for the ``concat'' baseline. This suggests that the first
6 layers do not contain useful for the test task information. For this reason,
we do not include them in the multi-domain feature set, when reporting the results
in Section 4.4.

We further provide analysis of selection coefficients assigned to different layers in
Figure~\ref{fig:lambdas_mini_imagenet}. We can see that for all methods, SUR
picks from the last 6 layers most of the time. However, it can happen that some
of the earlier layers are selected too. According to
Table~\ref{tab:mini_imagenet_all}, these cases lead to a decrease in performance
and suggest the SUR may overfit, when the number of samples if very low.

\end{document}